\documentclass{article}

\usepackage{arxiv}

\usepackage[utf8]{inputenc} 
\usepackage[T1]{fontenc}    
\usepackage{hyperref}       
\usepackage{url}            
\usepackage{booktabs}       
\usepackage{amsfonts}       
\usepackage{nicefrac}       
\usepackage{microtype}      
\usepackage{lipsum}
\usepackage{graphicx}
\graphicspath{ {./images/} }
\usepackage{graphicx}
\usepackage{amsmath}
\usepackage{booktabs}
\usepackage{microtype}
\usepackage{adjustbox}
\usepackage{comment}
\usepackage{eurosym}
\usepackage{amsmath}
\usepackage{amssymb}

\title{Reinforcement Learning for Sequential Solar PV Policy Design under Uncertainty: An Agent-Based Approach}

\author{
Iias Faiud$^{1}$ \quad
Jonaid Shianifar$^{1}$ \quad
Michael Schukat$^{1}$ \quad
Karl Mason$^{1}$
\\[1ex]
$^{1}$School of Computer Science, University of Galway \\
Galway, Ireland, H91 TK33 \\
Iias Faiud: \texttt{i.faiud1@universityofgalway.ie}
}

\begin{document}
\maketitle
\begin{abstract}
Designing effective and fiscally sustainable policies for solar photovoltaic (PV) adoption requires balancing adoption gains against public expenditure under uncertainty and heterogeneous decision-making. This study formulates PV policy design as a sequential decision problem and integrates reinforcement learning (RL) with a stochastic agent-based model (ABM) that simulates yearly solar PV adoption under uncertainty. A policymaker agent selects annual incentives, including capital grants, subsidised loan rates, and feed-in tariffs, over a 16-year horizon. Adoption--cost trade-offs are explored by varying policy preferences within a scalarised reward framework. Policies are learned using PPO, SAC, and TD3 and evaluated under stochastic simulation. The results show that this approach produces a clear trade-off structure: the highest-adoption policy (TD3, $w_{\text{cost}}=0.5$) achieves approximately 4,145 adopters at a cost of \euro41.73 million, while the lowest-cost policy (PPO, $w_{\text{cost}}=2.0$) reduces expenditure to \euro7.27 million with 2,682 adopters. The balanced policy (PPO, $w_{\text{cost}}=1.6$) achieves 3,495 adopters at a cost of \euro22.47 million. Across algorithms, consistent trade-off patterns are observed, indicating robustness of the adoption--cost relationship. Compared with static baseline policies, the RL framework explores a broader range of policy configurations. These findings demonstrate the potential of RL as a flexible tool for adaptive policy design under uncertainty.

\keywords{Reinforcement learning \and Agent-based modelling \and Solar photovoltaic \and Policy optimisation \and Sequential decision-making \and Energy policy}

\end{abstract}


\section{Introduction}

Accelerating the uptake of distributed renewable technologies such as solar photovoltaic (PV) systems is central to low-carbon energy transitions, yet adoption often remains below technical and economic potential. Prior research shows that PV uptake depends not only on financial returns but also on actor heterogeneity, behavioural factors, and social influences, making policy design more complex than a purely cost-driven problem \cite{rai2015agent,shakeel2023solar,jacksohn2019drivers,robinson2015determinants}. 

Two common approaches to PV policy analysis are scenario-based simulation and optimisation. Scenario-based simulation, particularly agent-based models (ABMs), captures heterogeneous adopters and interaction effects in adoption dynamics \cite{rai2015agent,zhang2016data,macal2005tutorial}. Optimisation approaches identify trade-offs among competing objectives, with multi-objective methods commonly used to recover trade-off surfaces \cite{delgarm2016novel,khezri2020review}. However, these approaches typically treat policy as static over the planning horizon.

In practice, policy design is inherently sequential, as incentives are adjusted in response to evolving costs, prices, and deployment outcomes. This motivates the use of reinforcement learning (RL), which enables learning decision policies through interaction with dynamic environments \cite{sutton1998reinforcement}. While widely applied in energy systems, RL has largely focused on operational control rather than adaptive policy design in socio-technical adoption settings \cite{vazquez2019reinforcement}.

This study addresses this gap by formulating PV policy design as a sequential decision problem and integrating RL with a stochastic ABM that simulates yearly PV adoption among Irish dairy farms under uncertainty. A policymaker agent selects annual policy instruments, including grant share, subsidised loan rate, and feed-in tariff, over a finite horizon, while the ABM captures heterogeneous adoption behaviour. A scalarised reward balances adoption and public cost, and varying the cost weight generates a family of policies approximating an adoption--cost trade-off frontier.

The framework is evaluated using PPO, SAC, and TD3 \cite{schulman2017proximal,haarnoja2018soft,fujimoto2018addressing}. Results show a consistent trade-off structure across algorithms: stronger incentives increase adoption but require substantially higher public expenditure. Compared with static baselines, the framework explores a broader range of dynamic policy configurations and provides a flexible approach to adaptive policy design under uncertainty.

This paper makes three contributions: (1) formulating PV policy design as a sequential decision problem, (2) integrating RL with a behaviourally grounded ABM, and (3) using scalarised RL to systematically explore adoption--cost trade-offs.

\section{Literature Review}

ABM is widely used to analyse complex systems where aggregate outcomes emerge from heterogeneous, interacting decision-makers, making it well suited to socio-technical transitions with diverse actors and feedback effects \cite{macal2005tutorial}. In solar PV adoption, ABMs represent uptake as a behavioural process rather than a purely economic response, integrating demographic, social, environmental, and economic factors to capture heterogeneous behaviour under uncertainty \cite{rai2015agent,zhang2016data,shakeel2023solar}.

Policy evaluation has also been examined using econometric and comparative approaches, particularly for feed-in tariffs (FiTs). Evidence from EU countries shows that FiT effectiveness depends strongly on policy design and market context rather than simple policy presence \cite{jenner2013assessing}. Practitioner-oriented analyses similarly highlight the importance of tariff structure, differentiation, and degression in shaping both deployment and public cost \cite{couture2010policymaker}, reinforcing that renewable energy policy is fundamentally a multi-dimensional design problem.

Optimisation approaches are widely used to analyse trade-offs among objectives such as cost, adoption, and sustainability. Multi-objective evolutionary algorithms (MOEAs), such as NSGA-II, are commonly applied due to their ability to recover non-dominated solutions in complex settings \cite{deb2002fast}. Reviews of energy system modelling confirm the increasing use of multi-objective optimisation to support trade-off analysis across economic, environmental, and social dimensions \cite{chen2023review}. However, these approaches typically treat policy as static rather than adaptive over time.

RL offers an alternative by learning policies in sequential, stochastic environments. In energy systems, RL has been widely applied to operational problems such as demand response \cite{vazquez2019reinforcement}, while recent work demonstrates the feasibility of integrating RL with agent-based simulation in domains such as electricity markets \cite{harder2025assume}. Related studies further show that public policy can be framed as a sequential learning problem, where agents learn dynamic interventions in simulated environments \cite{zheng2020ai}.

The relationship between multi-objective decision-making and RL requires careful interpretation because the RL agent optimises a scalarised reward for a given set of preference weights. Scalarisation provides a practical mechanism for combining competing objectives into a single reward while varying preference weights to explore trade-offs~\cite{roijers2013survey}. In this context, varying the cost weight enables the generation of a family of policies that approximate the adoption--cost trade-off.

Relative to this literature, this study formulates PV policy design as a sequential decision problem in which a policymaker learns adaptive policy trajectories within a stochastic ABM environment. This shifts the focus from static optimisation to dynamic policy design under uncertainty, while using scalarised RL as a practical tool for systematic trade-off exploration.

\section{Methodology}

\subsection{Problem Formulation}

This research formulates solar photovoltaic (PV) policy design as a finite-horizon Markov decision process (MDP), in which a policymaker agent learns sequential policy interventions within a stochastic agent-based simulation environment. The decision process unfolds over a horizon of $T = 16$ years, indexed by $t = 0, \dots, T-1$, where each time step corresponds to one year of policy implementation and system evolution.

\begin{figure}[t]
    \centering
    \includegraphics[width=\textwidth]{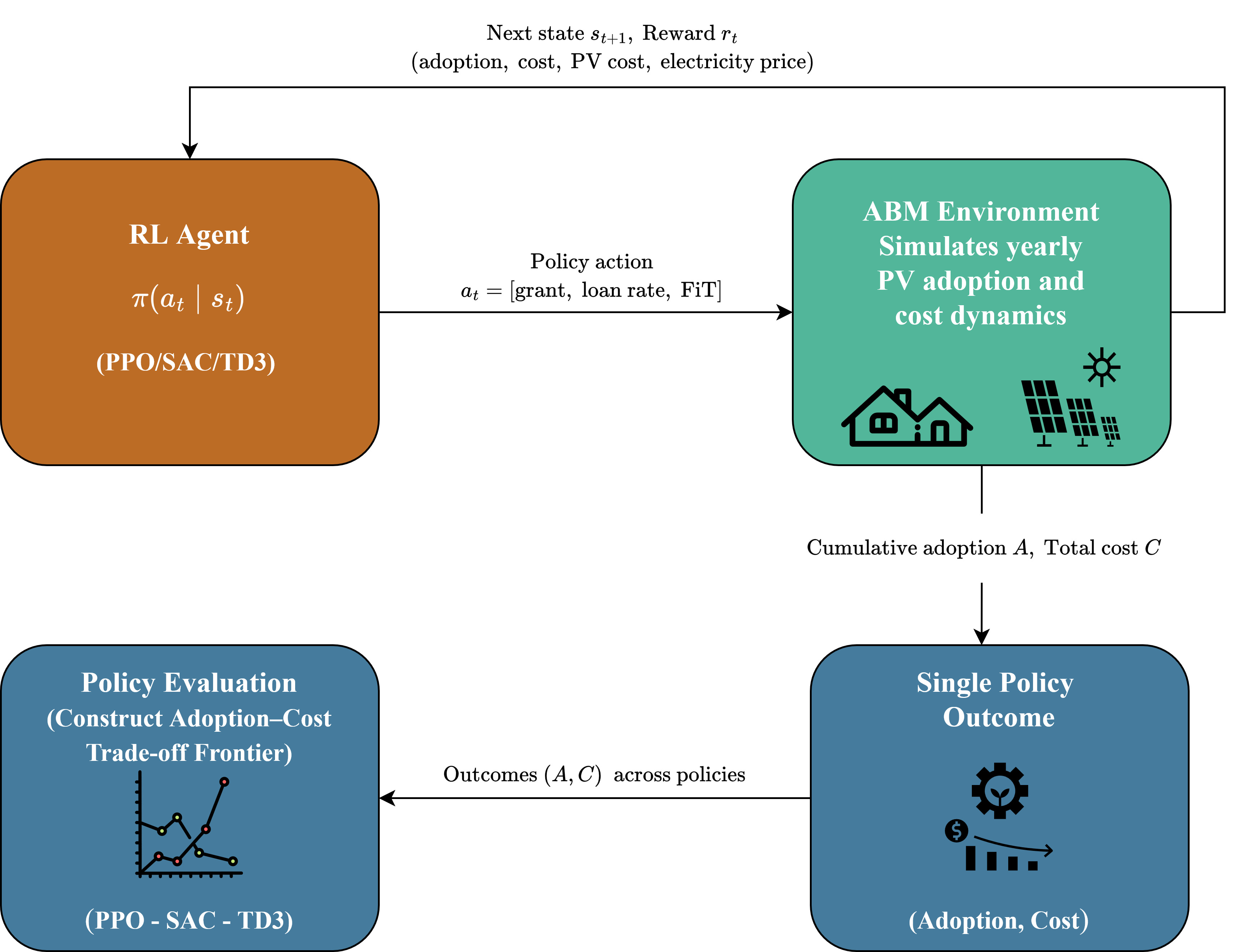}
    \caption{Sequential PV policy-design framework. The RL agent selects annual policy actions, which are evaluated by a stochastic ABM to produce adoption and cost outcomes used to construct the adoption--cost trade-off frontier.}

    \label{framework}
\end{figure}

Fig.~\ref{framework} illustrates the overall structure of the proposed framework. The policymaker, represented as an RL agent, selects policy actions based on the observed state. These actions are applied to the ABM, which simulates adoption dynamics under uncertainty and returns the resulting system state and reward. Policy outcomes, including cumulative adoption and public cost, are then used for offline evaluation to construct the adoption--cost trade-off frontier.

\paragraph{Observed state space.}

At each time step $t$, the agent observes a normalised vector $s_t \in \mathcal{S}$ capturing current system conditions, including the time index, cumulative and new adoption shares, PV cost level, electricity price level, cumulative public cost, and segment-level adoption rates across $K$ representative farm segments, where $K$ denotes the number of segments used to represent heterogeneity among dairy farms.

The observed state is:

\begin{equation}
s_t = \left[
\tilde{t}, \;
\frac{A_t}{N_t}, \;
\frac{\Delta A_t}{N_t}, \;
\tilde{c}^{\text{PV}}_t, \;
\tilde{p}^{\text{elec}}_t, \;
\tilde{C}_t, \;
r_{1,t}, \dots, r_{K,t}
\right]
\end{equation}

where $A_t$ is cumulative adoption, $\Delta A_t$ is new adoption in year $t$, $N_t$ is the total population in year $t$, $\tilde{c}^{\text{PV}}_t$ and $\tilde{p}^{\text{elec}}_t$ denote normalised PV cost and electricity price, $\tilde{C}_t$ is normalised cumulative public cost, and $r_{k,t}$ is the adoption rate in segment $k$. Variables marked with a tilde denote normalised quantities used for numerical stability and comparability across state variables.

The agent does not observe the full internal state of the simulation. Key variables such as scenario-specific uncertainty draws, export shares, and segment-level latent parameters are part of the simulator's internal dynamics but are not included in the observation.

\paragraph{Action space.}
At each time step, the agent selects a continuous policy action $a_t \in \mathcal{A}$ consisting of three instruments:
\begin{equation}
a_t = \left[ g_t, \; \ell_t, \; \tau_t \right]
\end{equation}
where $g_t$ is the grant share, $\ell_t$ is the subsidised loan rate, and $\tau_t$ is the feed-in tariff. The action space is continuous and bounded:
\begin{equation}
g_t \in [0, 0.6], \quad
\ell_t \in [0, 0.08], \quad
\tau_t \in [0.1, 0.3]
\end{equation}
Each action is applied for one year, after which the environment transitions to the next state.

These bounds are selected to reflect realistic policy design ranges consistent with existing and plausible renewable energy support schemes~\cite{FiT-ranges,Bank-of-ireland,Government-of-Ireland}, keeping the learned policies within predefined, policy-relevant ranges.

\paragraph{Transition dynamics.}
The transition function is defined by a stochastic ABM of PV adoption. Given state $s_t$ and action $a_t$, the environment simulates one year of system evolution, including updates to PV costs, electricity prices, and population characteristics.

Adoption decisions are generated probabilistically at the segment level. For each segment, a utility value is computed from techno-economic factors and policy incentives, and adoption probability is determined using a logistic function:
\begin{equation}
p_{k,t} = \frac{1}{1 + \exp\left( -(\alpha U_{k,t} + \beta) \right)}
\label{Logistic_Equation}
\end{equation}
where $U_{k,t}$ denotes the adoption utility for segment $k$ at time $t$, and $\alpha = 0.00005$ and $\beta = -6.34$ are calibrated behavioural parameters adopted from established models in the literature~\cite{faiud2024agent}.
New adopters are then sampled via a binomial process over the remaining non-adopters:
\begin{equation}
\Delta A_{k,t} \sim \text{Binomial}(N_{k,t}^{\text{rem}}, \, p_{k,t})
\end{equation}
where $N_{k,t}^{\text{rem}}$ is the number of remaining dairy farms in segment $k$. The total new adoption $\Delta A_t$ is obtained by aggregating across segments.

\paragraph{Reward function.}
The agent receives a scalar reward at each time step defined as:
\begin{equation}
R_t = w_{\text{adoption}} \cdot \hat{A}_t
- w_{\text{cost}} \cdot \hat{C}_t
- w_{\text{policy}} \cdot \delta_t
\end{equation}
where $\hat{A}_t$ denotes normalised new adoption in year $t$, $\hat{C}_t$ denotes normalised public cost incurred in that year, and $\delta_t$ is a policy adjustment penalty defined as the mean absolute change between consecutive normalised actions:
\begin{equation}
\delta_t = \frac{1}{d_a}\sum_{i=1}^{d_a}\left|\tilde{a}_t^{(i)}-\tilde{a}_{t-1}^{(i)}\right|
\end{equation}
where $d_a=3$ is the action dimension and $\tilde{a}_t^{(i)}$ denotes the bounds-normalised value of action component $i$, obtained by min--max scaling each action to $[0,1]$ using its lower and upper bounds. At the initial time step ($t=0$), the previous normalised action is initialised to zero, i.e., $\tilde{a}_{-1} = \mathbf{0}$.

The adoption and cost terms are normalised, and the weights $w_{\text{adoption}}$, $w_{\text{cost}}$, and $w_{\text{policy}}$ control the trade-off between adoption, public expenditure, and policy adjustment.

\paragraph{Objective.}
The policymaker seeks to maximise the expected discounted return over the planning horizon:
\begin{equation}
J(\pi) = \mathbb{E}_{\pi} \left[ \sum_{t=0}^{T-1} \gamma^t R_t \right]
\end{equation}
where $\pi$ is the policy and $\gamma = 0.99$ is the RL discount factor. In addition, the ABM internally applies an economic discount rate to cost flows, reflecting the time value of public expenditure.

\paragraph{Trade-off exploration.}
To explore the trade-off between adoption and public cost, the cost weight $w_{\text{cost}}$ is varied across a predefined range. This induces a family of policies corresponding to different policy preferences, enabling systematic analysis of the adoption--cost trade-off.

\subsection{Agent-Based Environment}
The RL agent interacts with a stochastic ABM of solar PV adoption based on an established framework~\cite{faiud2024agent}. The model simulates annual adoption decisions among heterogeneous Irish dairy farms, updating PV costs, electricity prices, population size, and policy costs under each selected action. The population is represented by $K$ segments to capture heterogeneity while maintaining computational tractability. Uncertainty is incorporated through Monte Carlo sampling of key drivers, including electricity price growth, PV cost decline, and export share.

\paragraph{Model inputs and calibration data.}
The ABM is parameterised using empirically grounded inputs, including electricity consumption levels, PV system costs~\cite{commercial-solar-panels-cost}, electricity tariffs~\cite{SEAI-Electricity-cost}, and financial parameters such as discount rates and maintenance costs~\cite{Maintenance-Cost-Solar-Panel}. The underlying adoption model was calibrated and evaluated in prior work against historical PV uptake among Irish dairy farms, demonstrating its ability to reproduce plausible adoption dynamics~\cite{faiud2024agent}. These inputs ensure consistency with observed market conditions and provide a data-informed basis for simulating adoption dynamics under alternative policy configurations.

\subsection{Reinforcement Learning Configuration}

The policymaker is modelled as an agent learning a policy $\pi(a_t \mid s_t)$ to maximise the expected discounted return defined in Section~3.1. Three continuous-control algorithms are used: PPO, SAC, and TD3, selected for their suitability to continuous action spaces. PPO is an on-policy actor--critic method based on clipped policy updates, while SAC and TD3 are off-policy actor--critic methods. SAC uses entropy regularisation to encourage exploration, whereas TD3 addresses overestimation bias through delayed policy updates and twin critics.

For each algorithm and each value of $w_{\text{cost}}$, a separate policy is trained for $1{,}000{,}000$ timesteps using five random seeds. Each episode corresponds to one complete 16-year policy trajectory, with annual time steps. Training uses a reduced Monte Carlo setting for computational efficiency, while final evaluation is performed on 300 independent stochastic episodes sampled from the same uncertainty distributions using higher-fidelity sampling. Reported adoption and cost outcomes are based on these final evaluation episodes.

To explore the adoption--cost trade-off, $w_{\text{cost}}$ is varied across a predefined range, producing a family of policies corresponding to different preference settings. Three static baseline policies representing low, moderate, and high support levels are evaluated under the same protocol and compared in terms of cumulative adoption, total public cost, and cost per adopter.

\section{Results}

\subsection{Training Behaviour}

\begin{figure}[t]
    \centering
    \includegraphics[width=\textwidth]{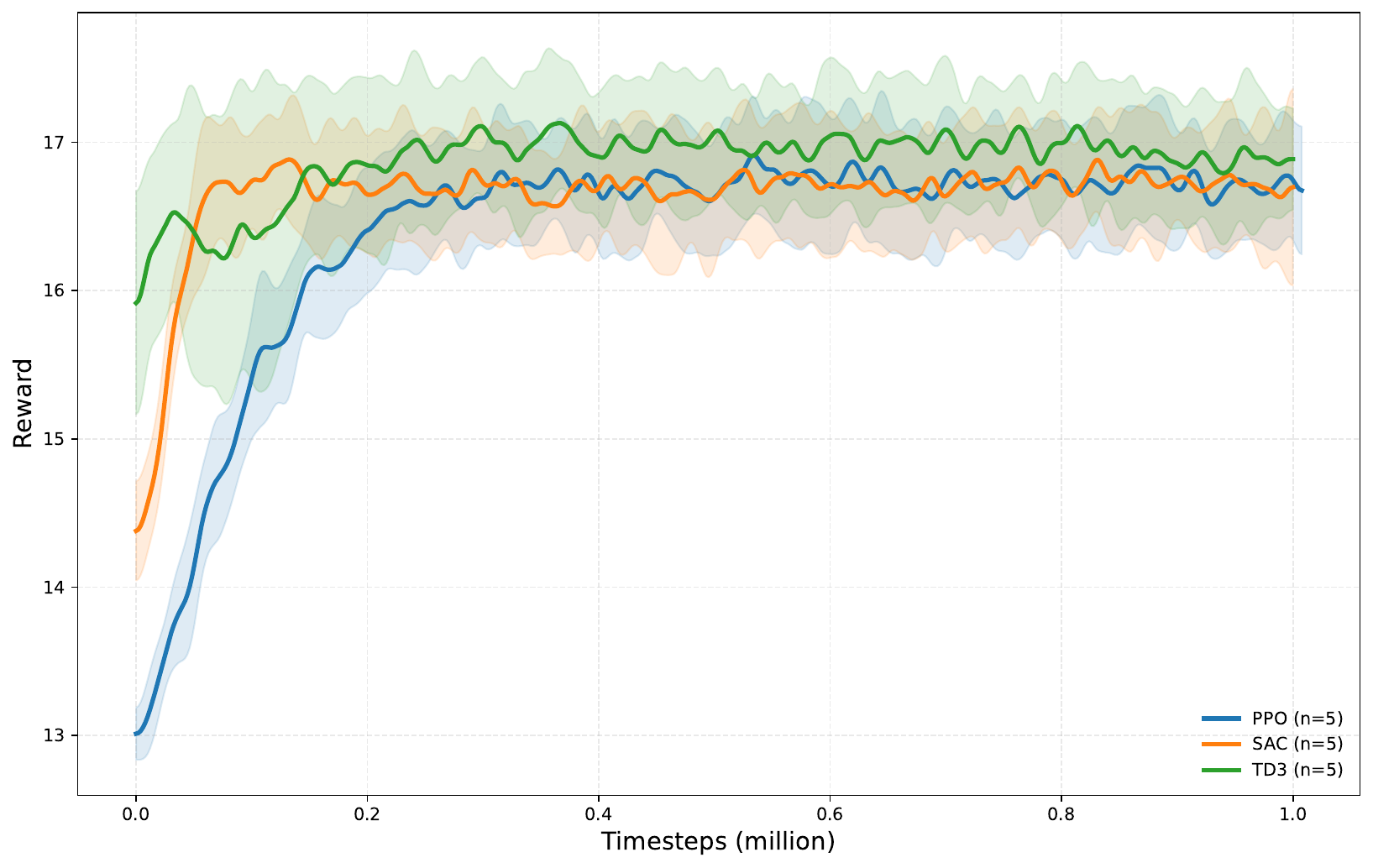}
    \caption{Training reward trajectories for PPO, SAC, and TD3 under $w_{\text{cost}}=0.5$. Lines show mean reward across five seeds and shaded regions show variability.}
    \label{training_curves}
\end{figure}

The training dynamics of PPO, SAC, and TD3 are illustrated in Fig.~\ref{training_curves} for a representative cost-weight configuration ($w_{\text{cost}}=0.5$). All three algorithms exhibit stable learning behaviour, with rewards increasing rapidly during early training and stabilising thereafter. Although variability remains due to stochastic simulation, the trajectories indicate consistent convergence across seeds.

No evidence of divergence is observed, suggesting that all three algorithms are well suited to this continuous, stochastic optimisation setting.

\subsection{Adoption--Cost Trade-off}

\begin{figure}[t]
    \centering
    \includegraphics[width=\textwidth]{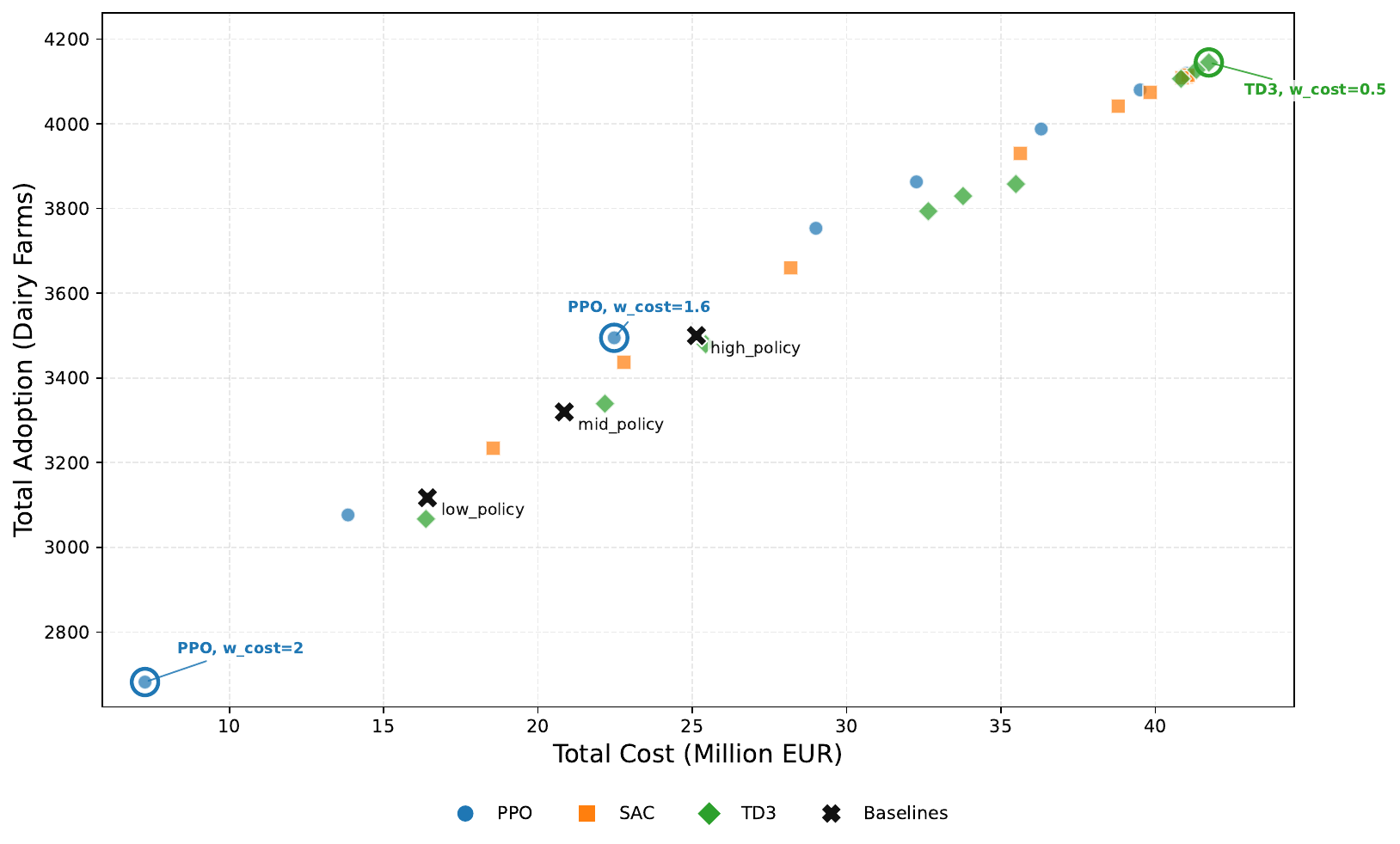}
    \caption{Adoption--cost trade-off across PPO, SAC, and TD3. Each point represents a learned policy under a specific cost-weight setting. Highlighted points indicate the representative policies used for baseline comparison.}
    \label{pareto_alg}
\end{figure}

The adoption--cost trade-off is shown in Fig.~\ref{pareto_alg}. A clear structure emerges across all algorithms: higher adoption requires substantially greater public expenditure. Representative policies are highlighted in the figure.

At the low-cost end, PPO with $w_{\text{cost}}=2.0$ yields approximately 2,682 adopters at a cost of about \euro7.27 million. At the opposite extreme, the highest-adoption configuration (TD3, $w_{\text{cost}}=0.5$) reaches approximately 4,145 adopters at about \euro41.73 million.

This pattern shows that higher adoption requires higher public expenditure, with the cost per adopter increasing for the highest-adoption policies. The consistency of the frontier across PPO, SAC, and TD3 indicates that the trade-off is driven primarily by system dynamics rather than algorithm-specific behaviour.

\subsection{Representative Policies}

To support interpretation, three outcome-based representative policies are selected from the pooled frontier obtained across PPO, SAC, and TD3: the low-cost policy (PPO, $w_{\text{cost}}=2.0$), the high-adoption policy (TD3, $w_{\text{cost}}=0.5$), and a balanced policy (PPO, $w_{\text{cost}}=1.6$). The algorithm associated with each point reflects the trained policy occupying that region of the frontier. The comparison therefore focuses on policy outcomes rather than algorithm ranking.

The balanced policy achieves approximately 3,495 adopters at a total cost of about \euro22.47 million. Compared with the moderate baseline, this represents a modest increase in adoption with only a moderate increase in public cost, while remaining well below the expenditure level of the high-adoption regime.

\subsection{Comparison with Baselines}

Three static baseline policies are considered, representing low, moderate, and high levels of policy support, each defined by fixed incentive levels over time.

Compared with the low-support baseline, the RL-derived low-cost policy (PPO, $w_{\text{cost}}=2.0$) reduces total public expenditure by 55.7\%, with a 14.0\% decrease in adoption and a 48.6\% reduction in cost per adopter. The balanced policy (PPO, $w_{\text{cost}}=1.6$) increases adoption by 5.3\% relative to the moderate baseline, with a 7.8\% increase in cost and a 2.5\% increase in cost per adopter. The high-adoption policy (TD3, $w_{\text{cost}}=0.5$) increases adoption by 18.4\% relative to the high-support baseline, with cost increasing by 66.1\% and cost per adopter by 40.0\%.

These comparisons do not identify a single universally best policy. Rather, they show that the RL framework can generate different adoption--cost trade-offs: lower-cost policies reduce expenditure at the expense of adoption, while higher-adoption policies require greater public spending. The preferred policy therefore depends on the policymaker's relative priorities for uptake and fiscal restraint.

\subsection{Policy Dynamics and RL--ABM Interaction}

\begin{figure}[t]
    \centering
    \includegraphics[width=\textwidth]{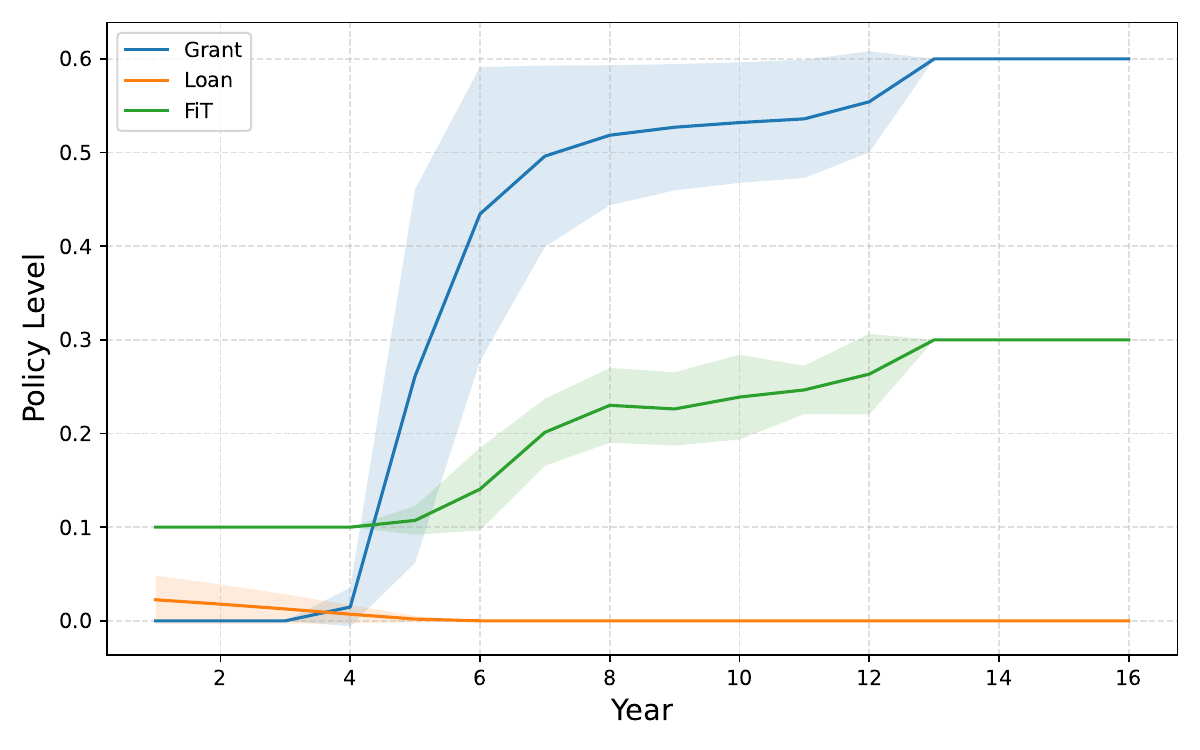}
    \caption{Policy trajectories for the balanced RL policy (PPO, $w_{\text{cost}}=1.6$). Shaded regions indicate variability across runs.}
    \label{ppoActions}
\end{figure}

To examine the interaction between the RL agent and the ABM, the evolution of policy actions over time was analysed for the balanced policy. Fig.~\ref{ppoActions} illustrates the trajectories for PPO with $w_{\text{cost}}=1.6$.

The learned policy evolves from low early support to stronger mid-horizon incentives. In particular, the grant share increases sharply after the early years and eventually reaches its upper bound, while the feed-in tariff rises more gradually and stabilises at its maximum level. By contrast, the loan rate declines towards zero.

This pattern indicates that the RL agent adjusts policy intensity in response to the adoption dynamics simulated by the ABM, increasing support when stronger incentives are needed and stabilising policy levels once adoption increases.

\section{Discussion and Conclusion}

The results show that solar PV policy design can be formulated as a sequential decision problem, enabling dynamic policy adaptation and more flexible exploration of adoption--cost trade-offs than static policy configurations.

A clear trade-off emerges between adoption and public expenditure: stronger incentives increase adoption, but at higher fiscal cost. The highest-adoption policies also involve higher cost per adopter. The consistency of this structure across PPO, SAC, and TD3 suggests that the results are driven mainly by system dynamics rather than algorithm-specific behaviour. Final policy outcomes are averaged over independent stochastic evaluation episodes, providing more stable estimates of expected adoption and public cost under uncertainty.

Several limitations remain. The study covers a 16-year horizon (2025--2040), one context, and a limited policy set. Although calibrated in prior work, the ABM simplifies adoption behaviour and omits some behavioural, financial, and implementation constraints. Policies reaching action bounds should be interpreted within the specified policy space. Future work could test alternative bounds, budgets, and implementation assumptions, transfer the RL--ABM framework to other contexts after recalibration, and compare it with multi-objective optimisation under consistent assumptions.

The proposed approach provides a practical framework for adaptive policy design in complex socio-technical systems, enabling systematic evaluation of policy trade-offs under uncertainty. In this context, the resulting policies should be interpreted as model-based scenarios for comparative policy analysis rather than as definitive policy recommendations.

\section*{Acknowledgements}
This publication has emanated from research conducted with the financial support of Research Ireland under Grant number [21/FFP-A/9040].

\bibliographystyle{unsrt}  
\bibliography{references}  

\end{document}